\documentclass[runningheads]{llncs}
\usepackage{graphicx}
\usepackage{amsmath,amssymb} 
\usepackage{color}

\usepackage{tabularx}
\usepackage{float}
\newfloat{figtab}{htb}{fgtb}
\makeatletter
  \newcommand\figcaption{\def\@captype{figure}\caption}
  \newcommand\tabcaption{\def\@captype{table}\caption}
\makeatother
\usepackage{multirow}
\usepackage{wrapfig}

\begin{document}
\newcolumntype{L}[1]{>{\raggedright\arraybackslash}p{#1}}
\newcolumntype{C}[1]{>{\centering\arraybackslash}p{#1}}
\newcolumntype{R}[1]{>{\raggedleft\arraybackslash}p{#1}}
\makeatletter
\DeclareRobustCommand\onedot{\futurelet\@let@token\@onedot}
\def\@onedot{\ifx\@let@token.\else.\null\fi}

\def\eg{\emph{e.g}\onedot} \def\Eg{\emph{E.g}\onedot}
\def\ie{\emph{i.e}\onedot} \def\Ie{\emph{I.e}\onedot}
\def\cf{\emph{c.f}\onedot} \def\Cf{\emph{C.f}\onedot}
\def\etc{\emph{etc}\onedot} \def\vs{\emph{vs}\onedot}
\def\wrt{w.r.t\onedot} \def\dof{d.o.f\onedot}
\def\etal{\emph{et al}\onedot}
%
\title{Dynamic Filtering with Large Sampling Field for ConvNets} 

\titlerunning{LS-DFN}
%
\author{Jialin Wu\thanks{Equal contribution}\inst{1,2}\orcidID{0000-0003-4684-5212} \and 
Dai Li$^\star$\inst{1} \and 
Yu Yang$^\star$\inst{1} \and Chandrajit Bajaj\inst{2} \and Xiangyang Ji\inst{1}}
%
\authorrunning{Wu et al}
%

\institute{The Department of Automation and BRNist, Tsinghua University, Beijing, 100084, China\\
\email{\{lidai15, yang-yu16\}@mails.tsinghua.edu.cn} \\
\email{xyji@tsinghua.edu.cn} \and The University of Texas at Austin, Austin TX 78712, USA\\
\email{\{jialinwu, bajaj\}@cs.utexas.edu} 
}
\maketitle              
\begin{abstract}
We propose a dynamic filtering strategy with large sampling field for ConvNets (LS-DFN), where the position-specific kernels learn from not only the identical position but also multiple sampled neighbour regions. During sampling, residual learning is introduced to ease training and an attention mechanism is applied to fuse features from different samples. Such multiple samples enlarge the kernels’ receptive fields significantly without requiring more parameters. While LS-DFN inherits the advantages of DFN \cite{de2016dynamic}, namely avoiding feature map blurring by positionwise kernels while keeping translation invariance, it also efficiently alleviates the overfitting issue caused by much more parameters than normal CNNs. Our model is efficient and can be trained end-to-end via standard back-propagation. 
We demonstrate the merits of our LS-DFN on both sparse and dense prediction tasks involving object detection, semantic segmentation and flow estimation. Our results show LS-DFN enjoys stronger recognition abilities in object detection and semantic segmentation tasks on VOC benchmark \cite{Everingham10} and sharper responses in flow estimation on FlyingChairs dataset \cite{DFIB15} compared to strong baselines.

\keywords{large sampling field, object detection, semantic segmentation, flow estimation}
\end{abstract}
\section{Introduction}
Convolutional Neural Networks have recently made significant progress in both sparse prediction tasks including image classification \cite{NIPS2012_4824,He_2016_CVPR,wang2017residual}, 
object detection \cite{dai2016r,ren2015faster,Girshick_2015_ICCV} and dense prediction tasks such as semantic segmentation  \cite{Long_2015_CVPR,dai2016instance,li2016fully}, flow estimation \cite{dosovitskiy2015flownet,ilg2016flownet,sun2017pwc}, \etc. 
Generally, deeper \cite{simonyan2014very,szegedy2015going,He_2016_CVPR} architectures provide richer features due to more trainable parameters and larger receptive fields. 

Most neural network architectures mainly adopt spatially shared kernels which work well in general cases. 
However, during training process, the gradients at each spatial position may not share the same descend direction, which can minimize loss at each position. These phenomena are quite ubiquitous when multiple objects appear in a single image in object detection or multiple object with different motion direction in flow estimation, which make the spatially shared kernels more likely to produce blurred feature maps.\footnote{Please see the examples and detailed analysis in the Supplementary Material.} The reason is that even though the kernels are far from optimal for every position, the global gradients, which are the spatially summation of the gradients over entire feature maps, can be close to zero. Because they are used in the update process, the back-propagation process should nearly not make progress.

Adopting position-specific kernels can alleviate the unshareable descend direction issue and take advantage of the gradients at each position (\ie\ local gradients) since kernel parameters are not spatially shared. In order to keep the translation invariance, Brabandere \etal\  \cite{de2016dynamic} propose a general paradigm called Dynamic Filter Networks (DFN) and verify them on moving MNIST dataset \cite{srivastava2015unsupervised}. 
However, DFN \cite{de2016dynamic} only generates the dynamic position-specific kernels for their own positions. As a result, the kernels can only receive the gradients from the identical position ($i.e.$ square of kernel size), which is usually more unstable, noisy and harder to converge than normal CNN.

\begin{figure}[!t]
\centering
\includegraphics[width=0.8\linewidth,trim={1.5cm 10.5cm 1.5cm 1.5cm},clip]{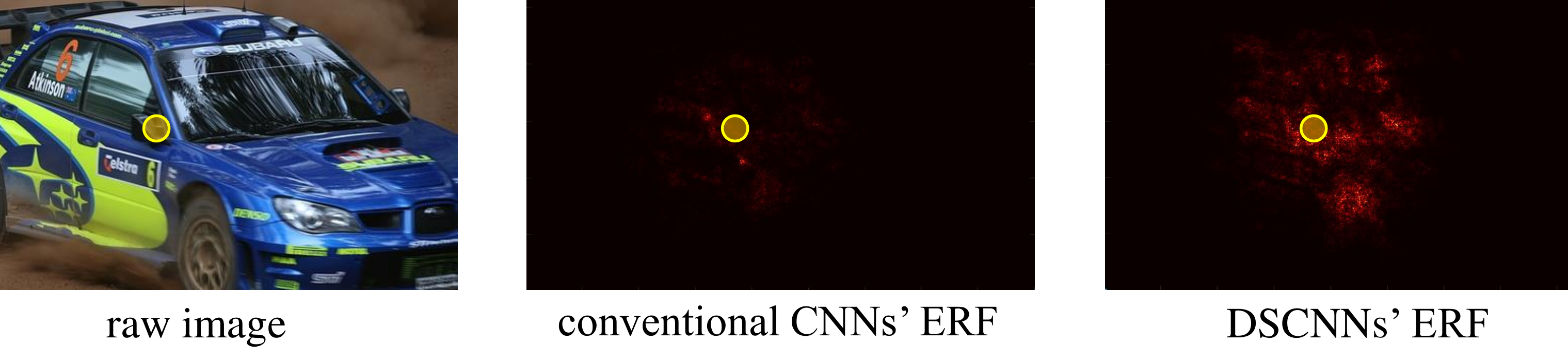}
\caption{Visualization of the effective receptive field (ERF). Yellow circle denotes the position on the object and the red region denotes the corresponding ERF.}
\label{fig:erf_demo}
\end{figure}
Meanwhile, properly enlarging receptive field is one of the most important concerns when designing CNN architectures. 
In many neural network architectures, adopting stacked convolutional layers with small kernels (\ie\ $3\times 3$) \cite{simonyan2014very} is more preferable than larger kernels (\ie\ $7\times 7$) \cite{NIPS2012_4824}, because the former one obtains the same receptive fields with fewer parameters. 
However, it has been shown that the effective receptive fields (ERF) \cite{Luo2016UnderstandingTE} only occupies a fraction of the full theoretical receptive field due to some weak connections and some unactivated ReLU units. 
In practice, it has been shown that adopting dilation strategies \cite{chen2016deeplab} can further improve performance \cite{dai2016r,li2016fully}, which means that enlarging receptive fields in a single layer is still beneficial. 

Therefore, we propose LS-DFN to alleviate the unshareable descend direction problem by utilizing dynamic position-specific kernels, and to enlarge the limited ERF by dynamic sampling convolution. 
As shown in Fig. \ref{fig:erf_demo}, with ResNet-50 as pretrained model, adding a single LS-DFN layer can significantly enlarge the ERF, which further results in the improvement on representation abilities. 
On the other hand, since our kernels at each position are dynamically generated, LS-DFNs also benefit from the local gradients. 
We evaluate our LS-DFNs via object detection and semantic segmentation tasks on VOC benchmark \cite{Everingham10} and optical flow estimation on FlyingChairs dataset \cite{DFIB15}. 
The results indicate that the LS-DFNs are general and beneficial for both sparse and dense prediction tasks. 
We observe improvements over strong baseline models in both tasks without heavy burden in terms of running time using GPUs.

\section{Related Work}
\noindent\textbf{Dynamic Filter Networks.}
Dynamic Filter Networks \cite{de2016dynamic} are originally proposed by Brabandere \etal\ to provide custom parameters for different input data. 
This architecture is powerful and more flexible since the kernels are dynamically conditioned on inputs. 
Recently, several task-oriented objectives and extensions have been developed.
Deformable convolution \cite{dai2017deformable} can be seen as an extension of DFNs that discovers geometric-invariant features.
Segmentation-aware convolution \cite{harley2017segmentation} explicitly takes advantage of prior segmentation information to refine feature boundaries via attention masks. Different from the models mentioned above, our LS-DFNs aim at constructing large receptive fields and receiving local gradients to produce sharper and more semantic feature maps. \\

\noindent\textbf{Receptive Field.}
Wenjie \etal propose the concept of effective receptive field (ERF) and the mathematical measure using partial derivatives. 
The experimental results verify that the ERF usually occupies only a small fraction of the theoretical receptive field \cite{Luo2016UnderstandingTE} which is the input region that an output unit depends on. Therefore, this has attracted lots of research especially in deep learning based computer vision. For instance, Chen \etal\ \cite{chen2016deeplab} propose dilated convolution with hole algorithm and achieve better results on semantic segmentation. 
Dai \etal\ \cite{dai2017deformable} propose to dynamically learn the spatial offset of the kernels at each position so that those kernels can observe wider regions in the bottom layer with irregular shapes. 
However, some applications such as large motion estimation and large object detection even require larger ERF.\\

\noindent\textbf{Residual Learning.}
Generally, residual learning reduces the difficulties of directly learning the objectives by learning their residual discrepancy of an identity function. 
ResNets \cite{He_2016_CVPR} are proposed to learn residual features of identity mapping via short-cut connection and helps deepen CNNs to over 100 layers easily. 
There have been plenty of works adopting residual learning to alleviate the problem of divergence and generate richer features. 
Kim \etal\ \cite{kim2016multimodal} adopt residual learning to model multimodal data in visual QA. 
Long \etal\ \cite{long2016unsupervised}learn residual transfer networks for domain adaptation. 
Besides, Fei Wang \etal\ \cite{wang2017residual} apply residual learning to alleviate the problem of repeated features in attention model. 
We apply residual learning strategy to learn residual discrepancy for identical convolutional kernels. 
By doing so, we can ensure valid gradients' back-propagation so that the LS-DFNs can easily converge in real-world datasets.\\

\noindent\textbf{Attention Mechanism.}
For the purpose of recognizing important features in deep learning unsupervisedly, attention mechanism has been applied to lots of vision tasks including image classification \cite{wang2017residual}, semantic segmentation \cite{harley2017segmentation}, action recognition \cite{sharma2015action,wu2016action}, \etc. In soft attention mechanisms \cite{sharma2015action,xu2015show,wang2017residual}, weights are generated to identify the important parts from different features using prior information. 
Sharma \etal\ \cite{sharma2015action} use previous states in LSTMs as prior information to have the network focus on more meaningful contents in the next frame and get better results for action recognition. Fei Wang \etal\ \cite{wang2017residual} benefit from lower-level features and learn attention for higher-level feature maps in a residual manner. In contrast, our attention mechanism aims at combining features from multiple samples via learning weights for each positions' kernels at each sample.

\section{Largely Sampled Dynamic Filtering}
Firstly, we present the overall structure of our LS-DFN in Sec. \ref{sec:3.1}, then introduce largely sampling strategies in Sec. \ref{sec:3.2}. This design allows kernels at each position to take advantage of larger receptive fields and local gradients. 
Furthermore, attention mechanisms are utilized to enhance the performance of LS-DFNs as demonstrated in Sec. \ref{sec:3.3}.
Finally, Sec. \ref{sec:3.4} explains implementation details of our LS-DFNs, $i.e.$ parameters reducing and residual learning techniques.
\subsection{Network Overview}
\label{sec:3.1}
\begin{figure}[!t]
\centering
\includegraphics[width=0.9\linewidth,height=5cm,trim={1.5cm 1.6cm 1.5cm 1.5cm},clip]{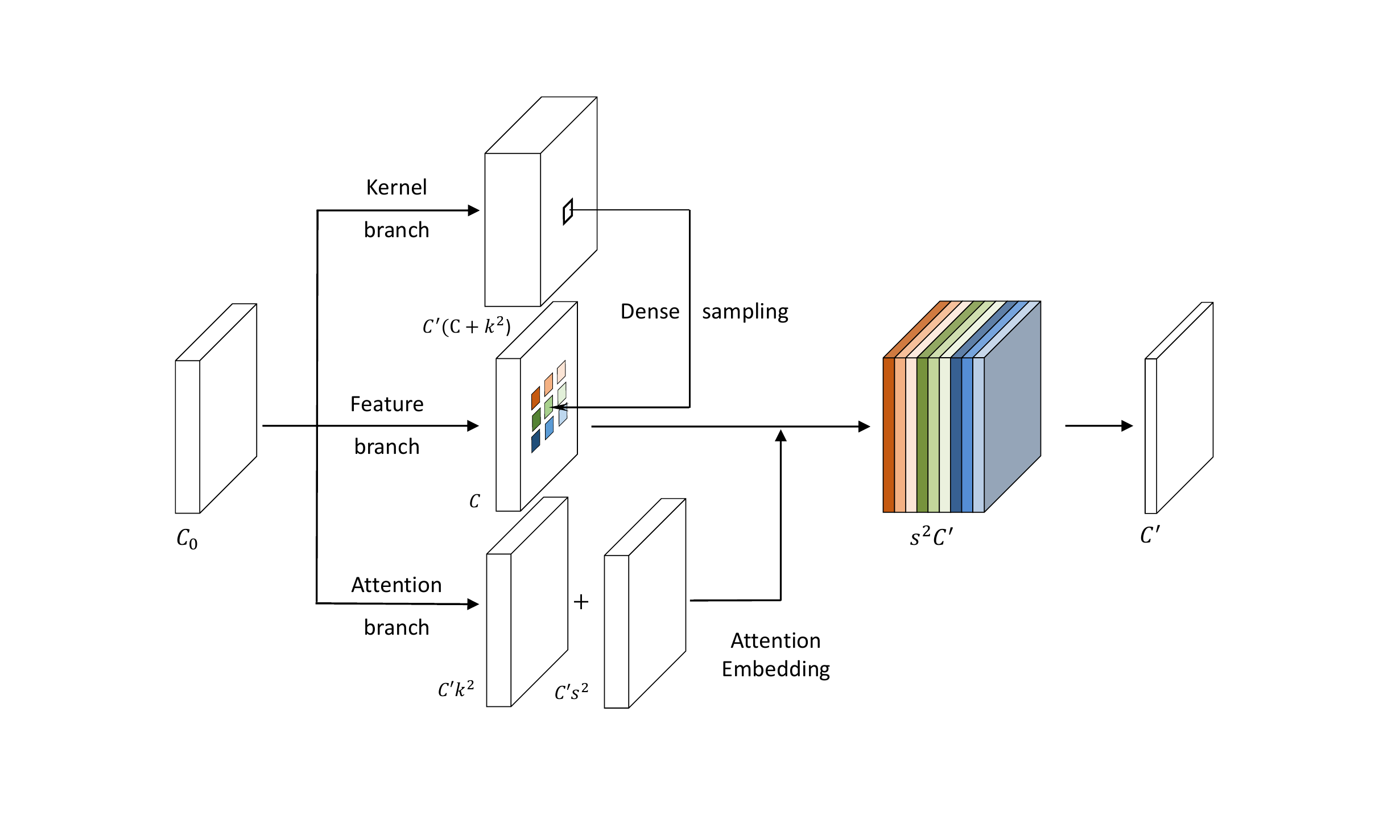}
\caption{Overview of the LS-DFN block. Our model consists of three branches: (1) the kernel branch generates position-specific kernels;
(2) the feature branch generates features to be position-specifically convolved;
(3) the attention branch generates attention weights. Same color indicates features correlated to the same spatial sampled regions.}
\label{fig:network_structure}
\end{figure}

We introduce the LS-DFNs' overall architecture in Fig. \ref{fig:network_structure}. Our LS-DFNs consist of three branches: (1) the feature branch firstly produces $C$ ($e.g. \ 128$) channels intermediate features; (2) the kernel branch, implemented as a convolution layers with $C'(C+k^2)$ channels where $k$ is kernel size, generates position-specific kernels to sample multiple neighbour regions in feature branches and produces $C'$ ($e.g. \ 32$) output channels' features; (3) the attention branch, implemented as convolution layers with $C'(s^2+k^2)$ channels where $s$ is the sampling size, outputs attention weights for each position's kernels and each sampling region. The LS-DFNs output feature maps with $C'$ channels and preserve the original spatial dimensions $H$ and $W$.

\subsection{Largely Sampled Dynamic Filtering}
\label{sec:3.2}
This subsection demonstrates the proposed largely sampled dynamic filtering enjoying both large receptive fields and the local gradients. In particular, the LS-DFNs firstly generate position-specific kernels by the kernel branch. After that, LS-DFNs further convolve these generated kernels with features from multiple neighbor regions in the feature branch to obtain large receptive fields. 

Denoting $\textbf{X}^{l}$ as the feature maps from $l^{th}$ layer(or intermediate features from feature branch) with shape $(C,H,W)$, normal convolutional layer with spatially shared kernels $\textbf{W}$ can be formulated as 
\begin{equation}
    \textbf{X}_{y,x}^{l+1,v} = \sum\limits_{u = 1}^{C}\sum\limits_{j = 0}^{k - 1}\sum\limits_{i = 0}^{k-1} \textbf{X}^{l,u}_{y+j,x+i}\textbf{W}^{v,u}_{y,x,j,i}
    \label{eq:norm_conv}
\end{equation}
where $u,v$ denote the indices of the input and output channels, $x,y$ denote the spatial coordinates and $k$ indicates the kernel size.

In contrast, the LS-DFNs treat generated features in kernel branch, which is spatially dependent, as convolutional kernels. 
This scheme requires the kernel branch to generate kernels $\mathcal{W}(X^l)$ from $X^{l}$, which can maps the $C$-channel features in the feature branch to $C'$-channel ones\footnote{$\mathcal{W}(X^l)$ is kernels generated from $X^l$, and we omit $(X^l)$ when there is no ambiguity.}. 
Detailed kernel generation methods will be described in Sec. \ref{sec:3.4} and the supplementary material.

\begin{wrapfigure}{R}{0.5\textwidth}
    \centering
    \includegraphics[width=\linewidth, trim={0cm 8cm 6cm 0cm}, clip]{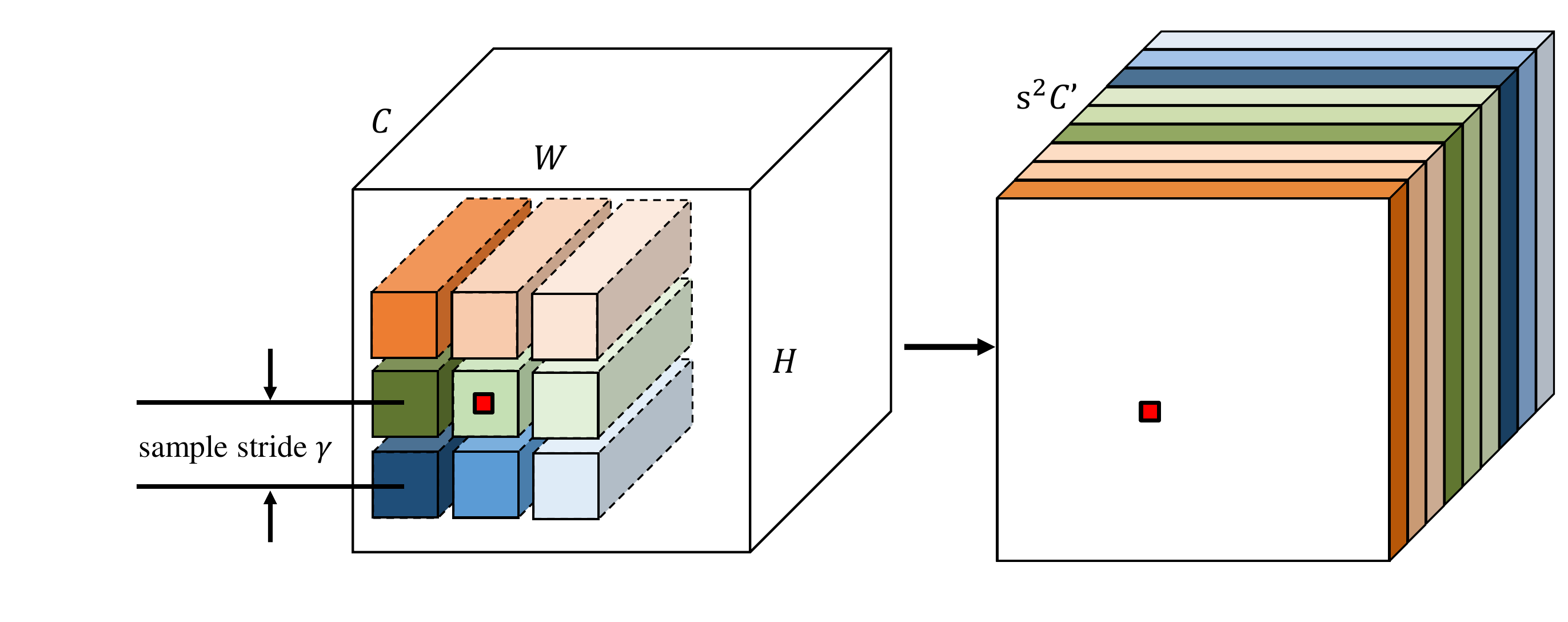}
\caption{Illustration of our sampling strategy. The red dot denotes the sampling point. Same color indicates features correlated to the same spatial sampled regions.}
\label{fig:dense_sample}
\end{wrapfigure}

Aiming at larger receptive fields and more stable gradients, we not only convolve the generated position-specific kernels with features at the identical positions in the feature branch, but also sample their $s^2$ neighbor regions as additional features as shown in Eq. \ref{eq:dynamic_sampling_conv}. Therefore, we have more learning samples for each position-specific kernel than DFN \cite{de2016dynamic}, resulting in more stable gradients. Also, since we obtain more diverse kernels (\ie\ position-specific) than conventional CNNs, we can robustly enrich the feature space. 

As shown in Fig. \ref{fig:dense_sample}, each position (\eg \ the red dot) outputs its own kernels in the kernel branch and uses the generated kernels to sample the corresponding multiple neighbour regions (\ie \ the cubes in different colors) in the feature branch. Assuming we have $s^2$ sampled regions for each position with sample stride $\gamma$, kernel size $k$, the sampling strategy outputs feature maps with shape $(s^2, C', H, W)$ which obtain approximately $(s\gamma)^2$ times larger receptive fields. 

Largely sampled dynamic filtering thus can be formulated as
\begin{equation}
    \hat{\textbf{X}}_{\alpha,\beta,y,x}^{l+1,v} = \sum\limits_{u = 1}^{C}\sum\limits_{i = 0}^{k-1}\sum\limits_{j = 0}^{k-1} \textbf{X}_{\hat{y}+j,\hat{x}+i}^{l,u}\mathcal{W}^{v,u}_{y,x,j,i},
    \label{eq:dynamic_sampling_conv}
\end{equation}\\
where $\hat{x}= x+\alpha\gamma$ and $\hat{y}= y+\beta\gamma$ denote the coordinates of the center in sampled neighbor regions. 
$\mathcal{W}$ denotes the position-specific kernels generated by the kernel branch. 
And $(\alpha, \beta)$ is the index of sampled region with sampling stride  $\gamma$. And when $s=1$, that LS-DFNs reduce to the origin DFN.

\subsection{Attention Mechanism}
\label{sec:3.3}
We present our methods to fuse features from multiple sampled regions at each position $\hat{\textbf{X}}_{\alpha,\beta,y,x}^{l+1,v}$. A direct solution is to stack $s^2$ sampled features to form a $(s^2C', H, W)$ tensor or perform a pooling operation on the sample dimension (\ie \ first dimension of $\hat{\textbf{X}}^{l+1}$) as outputs. However the first choice violates translation invariance and the second choice is not aware of which samples are more important.

To address this issue, we present an attention mechanism to fuse those features via learning attention weights for each position's kernel at each sample. Since the attention weights are also position-specific, the resolution of output feature maps can be potentially preserved. Also, our attention mechanism benefits from residual learning. 

Considering $s^2$ sampled regions and kernel size $k$ in each position, we should have $s^2 \times k^2 \times C'$ attention weights for each position for $\hat{\textbf{X}}^{l+1}$, which means
\begin{align}
\widetilde{\textbf{X}}_{\alpha,\beta,y,x}^{l+1,v} = 
\sum\limits_{u = 1}^{C} \sum\limits_{j = 0}^{k-1} \sum\limits_{i = 0}^{k-1} \textbf{X}_{\hat{y}+j,\hat{x}+i}^{l,u}\mathcal{W}^{v,u}_{y,x,j,i}\textbf{A}^{v,\alpha,\beta}_{\hat{y},\hat{x},j,i},
\label{eq:attention_origin}
\end{align}
where $\widetilde{\textbf{X}}$ denotes weighted features.
 
However, Eq. 3 requires $s^2k^2C'HW$ attention weights, which is computationally costly and easily leads to overfitting.
We thus split this task into learning position attention weights $\textbf{A}^{pos} \in \mathbb{R}^{k^2 \times C'\times H\times W}$ for kernels at each position and learning sampling attention weights $\textbf{A}^{sam} \in \mathbb{R}^{s^2\times C'\times H\times W}$ at each sampled region. 
Then Eq. \ref{eq:attention_origin} becomes
\begin{align}
 \widetilde{\textbf{X}}_{\alpha,\beta,y,x}^{l+1,v} = \textbf{A}^{sam,v}_{\alpha,\beta,y,x}\sum\limits_{u = 1}^{C}\sum\limits_{j = 0}^{k-1}\sum\limits_{i = 0}^{k-1} \textbf{X}_{\hat{y}+j,\hat{x}+i}^{l,u}\mathcal{W}^{v,u}_{y,x,j,i}\textbf{A}^{pos, v}_{\hat{y},\hat{x},j,i}
 ,
\label{eq:attention_separate}
\end{align}
where $\hat{y},\hat{x}$ share the same representations in Eq.\ref{eq:dynamic_sampling_conv}.

\begin{figure}[t]
\centering
\includegraphics[width=0.9\linewidth,trim={7cm 3cm 7cm 0cm},clip]{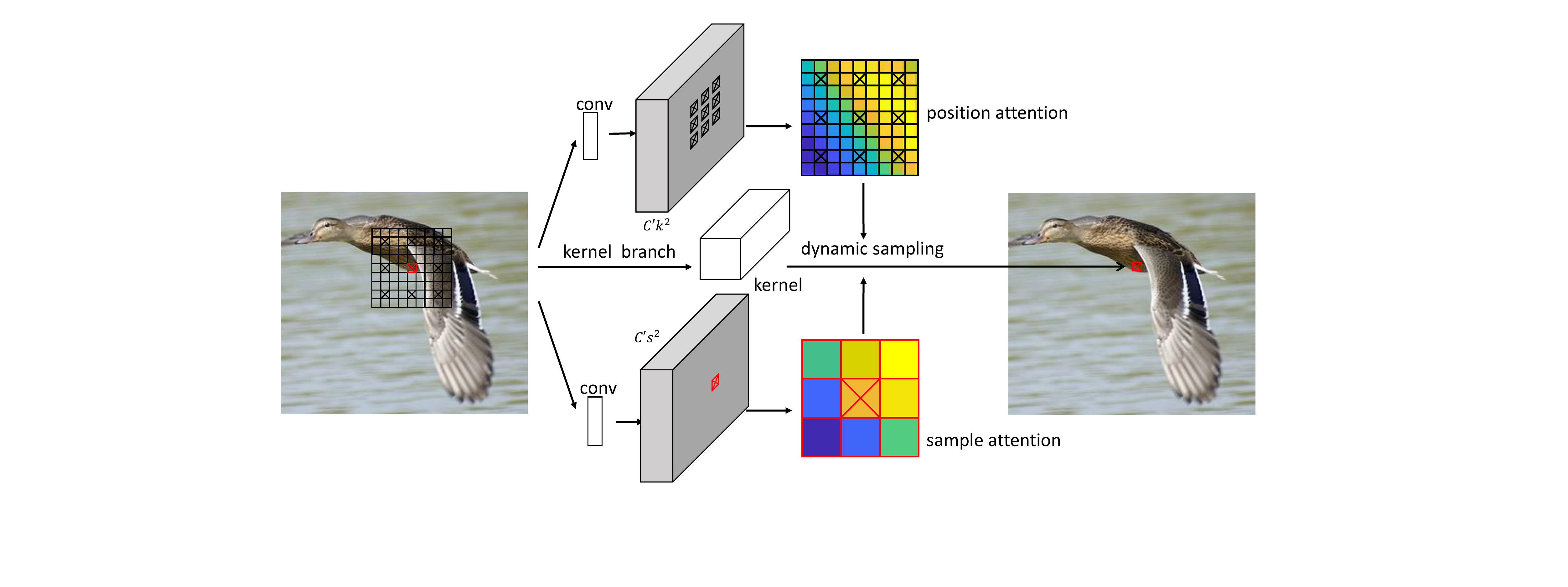}
\caption{At each position, we separately learn attention weights for each kernel and for each sample. Then, we combine features from multiple samples via these learned attention weights. Boxes with crosses denote the position to generate attention weights and red one denotes sampling position and black ones denote sampled positions.}
\label{fig:attention}
\end{figure}

Specifically, we use two CNN sub-branches to generate the attention weights for samples and positions respectively. 
The sampling attention sub-branch has $C' \times s^2 $ output channels and the position attention sub-branch has $C' \times k^2 $ output channels. 
The sample attention weights are generated from the sampling position denoted by the red box with cross in Fig.\ref{fig:attention} to coarsely predict the importance according to that position.
And the position attention weights are generated from each sampled regions denoted by black boxes with cross to model fine-grained local detailed importance based on the sampled local features. Further, we manually add $1$ to each attention weight to take advantage of residual learning.

Therefore, the number of attention weights will be reduced from $s^2k^2C'HW$ to $(s^2+k^2)C'HW$ as shown in Eq. \ref{eq:attention_separate}.
Obtaining Eq. \ref{eq:attention_separate}, we finally combine different samples via attention mechanism as 
\begin{equation}
    \textbf{X}_{y,x}^{l+1,v} = \sum\limits_{\alpha = 0}^{s-1}\sum\limits_{\beta = 0}^{s-1} \widetilde{\textbf{X}}_{\alpha,\beta,y,x}^{l+1,v}
    .
    \label{eq:attention_combine}
\end{equation}

Noting that feature maps from previous normal convolutional layers might still be noisy, the position attention weights help to filter such noise when applying largely sampled dynamic filtering to such feature maps. And the sample attention weights indicate how much contribution each neighbor region makes. 

\subsection{Dynamic Kernels Implementation Details}
\label{sec:3.4}
\noindent\textbf{Reducing Parameter.}
Given that directly generating the position-specific kernels $\mathcal{W}$ with shape same as conventional CNN will require the shape of the kernels to be $(C'Ck^2,H,W)$ as shown in Eq. \ref{eq:dynamic_sampling_conv}. 
Since $C$ and $C'$ can be relatively large (\eg\ up to 128 or 256), the required output channels in the kernel branch (\ie\ $C'Ck^2$) can easily get up to hundreds of thousands, which is computationally costly. Recently, several works have focused on reducing kernel parameters (\eg\ MobileNet \cite{Howard2017MobileNetsEC}) by factorizing kernels into different parts to make CNNs efficient in modern mobile devices. Inspired by them and based on our LS-DFNs' case, we describe our proposed parameter reduction method. And we provide the evaluation and comparison with state-of-art counterparts in the supplementary material. 

\begin{wrapfigure}{R}{0.5\textwidth}
    \centering
        \includegraphics[width=\linewidth, trim={0cm 12.5cm 13cm 0cm}, clip]{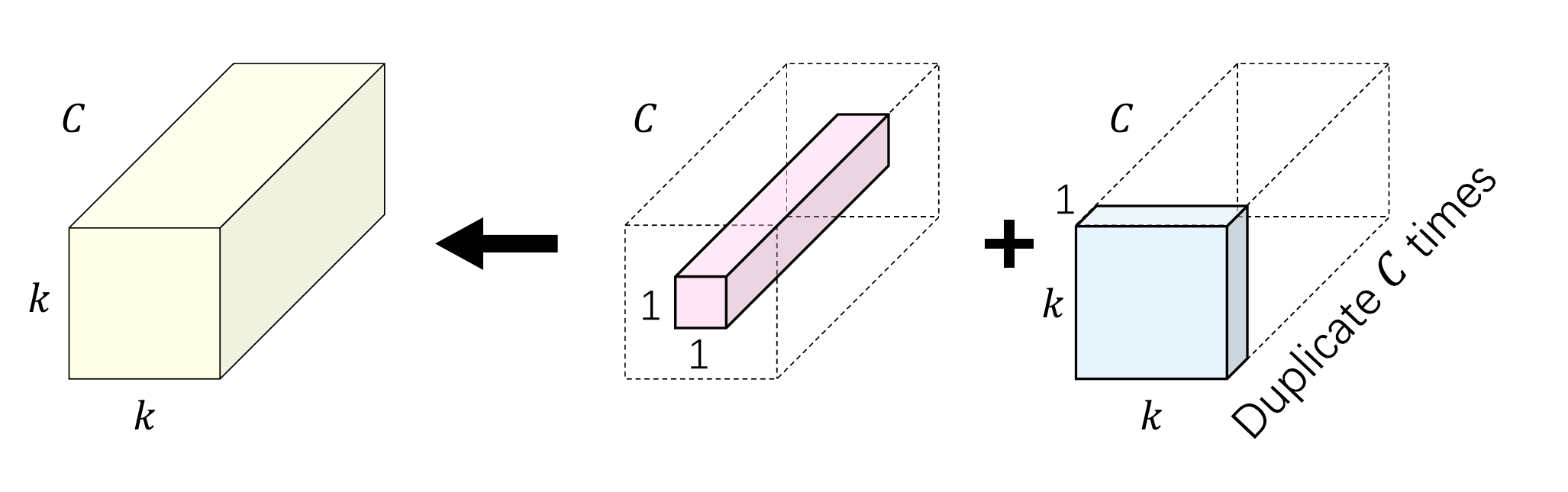}
    \caption{Illustration of our parameter reducing method. 
            In the first part, $C \times 1 \times 1$ weights are placed in the center of the corresponding kernel and in the second part $k^2$ weights are duplicated $C$ times.}
    \label{fig:kernel_reduce_param}
\end{wrapfigure}

Inspecting that activated output feature maps in a layer usually share similar geometric characteristics across channels, we propose a novel kernel structure that splits the original kernel into two separate parts for the purpose of parameter reduction. 
As illustrated in Fig. \ref{fig:kernel_reduce_param}, on the one hand, the $C \times 1 \times 1$ part $\mathcal{U}$ at each position, which will be placed into the spatial center of each $k \times k$ kernel, is used to model the difference across channels. On the other hand, the $1 \times k \times k$ part $\mathcal{V}$ at each position is used to model the shared geometric characteristics within each channel.

Combining the above two parts together, our method generates kernels that map $C$-channel feature maps to $C'$-channel ones with kernel size $k$ by only $C'(C+k^2)$ parameters at each position instead of $C'Ck^2$. 
Formally, the convolutional kernels used in Eq. \ref{eq:dynamic_sampling_conv} become 

\begin{align}
\mathcal{W}^{v,u}_{y,x,j,i} =
    \begin{cases}
      \mathcal{U}^{v,u}_{y,x} + \mathcal{V}^{v}_{y,x,j,i}  
      & \text{$j=i=\lfloor \frac{k-1}{2} \rfloor$}
    \\[4pt]
      \mathcal{U}^{v,u}_{y,x}   & \text{otherwise}
    \end{cases}.
    \label{eq:two_step_weight}
\end{align} \\

\noindent\textbf{Residual Learning.}
Eq. \ref{eq:two_step_weight} directly generates kernels, which easily leads to divergence in noisy real-world datasets. The reason is that only if the convolutional layers in kernel branch are well trained can we have good gradients back to feature branch and vice versa. Therefore, it's hard to train both of them from scratch simultaneously. Further, since kernels are not shared spatially, gradients at each position are more likely to be noisy, which makes kernel branch even harder to train and further hinders the training process of feature branch. 

We adopt residual learning to address this issue, which learns the residual discrepancies of identical convolutional kernels. In particular, we add $\frac{1}{C}$ to each central position of the kernels as
\begin{align}
\mathcal{W}^{v,u}_{y,x,j,i} =
    \begin{cases}
      \mathcal{U}^{v,u}_{y,x} + \mathcal{V}^{v}_{y,x,j,i} +
      \frac{1}{C}
      & \text{$j=i=\lfloor \frac{k-1}{2} \rfloor$}
    \\[4pt]
      \mathcal{U}^{v,u}_{y,x}   & \text{otherwise}
    \end{cases}.
    \label{eq:two_step_weight_residual}
\end{align}
Initially, since the outputs of the kernel branch are close to zero, LS-DFN approximately averages features from feature branch. 
It guarantees gradients are sufficient and reliable for back propagation to the feature branch, which inversely benefits the training process of the kernel branch. 

\section{Experiments}
We evaluate our LS-DFNs via object detection, semantic segmentation and optical flow estimation tasks. 
Our experiment results show that firstly with larger receptive fields, LS-DFN is more powerful on object recognition tasks. 
Secondly, with position-specific dynamic kernels and local gradients, LS-DFN produces much sharper optical flow.
Besides, the comparison between ERF of the LS-DFNs and conventional CNNs is also presented in Sec. \ref{sec:4.1}. 
This also verifies our aforementioned design target that LS-DFNs have larger ERF.

In the following subsections, we use $w/$ denotes with, $w/o$ denotes without, $\mathcal{A}$ denotes attention mechanism and $\mathcal{R}$ denotes residual learning, $C'$ denotes the number of dynamic features. Since $C'$ in our LS-DFN is relatively small ($e.g.$ 24) compared with conventional CNNs' settings, we optionally apply a post-conv layer to increase dimension to $C_1$ channels to match the conventional CNNs.

\subsection{Object Detection}
\label{sec:4.1}
We use \textit{PASCAL VOC} datasets \cite{Everingham10} for object detection tasks. 
Following the protocol in \cite{Girshick_2015_ICCV}, we train our LS-DFNs on the union of VOC 2007 trainval and VOC 2012 trainval and test on VOC 2007 and 2012 test sets. 
For evaluation, we use the standard mean average precision (mAP) scores with IoU thresholds at 0.5.

When applying our LS-DFN, we insert it into object detection networks such as R-FCN and CoupleNet. 
In particular, it is inserted right between the feature extractor and the detection head, producing $C'$ dynamic features. 
It is noting that these dynamic features just serve as complementary features, which are concatenated with original features before fed into detection head.
For R-FCN, we adopt ResNet as feature extractor and 7x7 bin R-FCN [7] with OHEM [32] as detection head. 
During training process, following \cite{dai2017deformable}, we resize images to have a shorter side of 600 pixels and adopt SGD optimizer. 
Following \cite{lin2016feature}, we use pre-trained and fixed RPN proposals.
Concretely, the RPN network is trained separately as in the first stage of the procedure in \cite{ren2015faster}. 
We train 110k iterations on single GPU with learning rate $10^{-3}$ in the first 80k and $10^{-4}$ in the next 30k. 

\begin{table}[!t]
\begin{minipage}[!t]{0.5\textwidth} 
    \centering
    \begin{tabular}{l|C{1.4cm}|C{1.4cm}}
    \hline
                                            & \scriptsize{mAP(\%) on VOC12} & \scriptsize{mAP(\%) on VOC07}  \\
    \hline\hline
    R-FCN \cite{dai2016r}                   & 77.6                          & 79.5              \\
    R-FCN+LS-DFN                            & \textbf{79.2}                 &\textbf{81.2}      \\
    \hline 
    Deform. Conv. \cite{dai2017deformable}  & -                             & 80.6              \\
    CoupleNet \cite{zhu2017couplenet}       & 80.4                          & 81.7              \\
    CoupleNet+LS-DFN                        & \textbf{81.7}$^\text{\dag}$   & \textbf{82.3}     \\
    \hline 
    \end{tabular}
    \caption{Evaluation of the LS-DFN models on VOC 2007 and 2012 detection dataset. We use $s=3$, $C'=24$, $\gamma = 1$, $C_1$ = 256 with ResNet-101 as pre-trained networks in experiments when adding LS-DFN layers. $^\text{\dag}$\url{http://host.robots.ox.ac.uk:8080/anony-mous/BBHLEL.html}.}
    \label{eval_voc_det}
\end{minipage}
\qquad
\begin{minipage}[!t]{0.43\textwidth} 
\centering
\begin{tabular}{L{2.2cm}|C{0.9cm}|C{0.9cm}|C{0.9cm}}
\hline
                                & $s=1$ & $s=3$ & $s=5$ \\
\hline\hline
$C'=16, w/ \mathcal{A}$   & 72.1  & 78.2  & 78.1 \\
$C'=24, w/ \mathcal{A}$   & 72.5  & 78.6  & 78.6 \\ 
$C'=32, w/ \mathcal{A}$   & 72.9  & 78.6  & 78.5 \\
\hline
\end{tabular}
\caption{Evaluation of numbers of samples $s$. The listed results are trained with residual learning and the post-conv layer is not applied. The experiments use R-FCN baseline and adopt ResNet-50 as pretrained networks.}
\label{eval_sample}
\end{minipage}
\begin{minipage}[!t]{0.5\textwidth} 
    \centering
    \begin{tabular}{C{1.2cm}|C{1.1cm}|C{1.1cm}|C{1.1cm}|C{1.1cm}}
    \hline
                & \multicolumn{2}{c|}{$\gamma=1$ } & \multicolumn{2}{c}{$\gamma=2$} \\
    \hline
                & $w/$ $\mathcal{A}$    & $w/o$ $\mathcal{A}$   & $w/$ $\mathcal{A}$    & $w/o$ $\mathcal{A}$  \\
    \hline\hline
    $C' = 16$   & 77.8                  & 77.4                  & 78.2                  & 77.4                 \\
    $C' = 24$   & 78.1                  & 77.4                  & 78.6                  & 77.3                 \\
    $C' = 32$   & 78.6                  & 77.6                  & 78.0                  & 77.3                 \\
    \hline
    \end{tabular}
    \caption{Evaluation of attention mechanism with different sample strides and numbers of dynamic features. The post-conv layer is not applied. The experiments use R-FCN baseline and adopt ResNet-50 as pretrained networks.}
    \label{eval_attention}
\end{minipage}
    \qquad
\begin{minipage}[!t]{0.42\textwidth} 
    \centering
    \begin{tabular}{C{1.2cm}|C{1.2cm}|C{1.1cm}|C{1.2cm}}
    \hline
    \multicolumn{2}{c|}{} & $w/$ $\mathcal{A}$    & $w/o$ $\mathcal{A}$     \\
    \hline
    \multirow{2}{*}{$C'=24$}   & $w/$ $\mathcal{R}$    & 78.68                 & 77.4                    \\
                                & $w/o$ $\mathcal{R}$   & 68.1                  & $\mathcal{F}$           \\
    \hline
    \multirow{2}{*}{$C'=32$}   & $w/$ $\mathcal{R}$    & 78.6                  & 77.6                    \\
                                & $w/o$ $\mathcal{R}$   & 68.7                  & $\mathcal{F}$           \\
    \hline
    \end{tabular}
    \caption{Evaluaion of residual learning strategy in LS-DFN. $\mathcal{F}$ indicates that the model fails to converge and the post-conv layer is not applied. The experiments use R-FCN baseline and adopt ResNet-50 as pretrained networks.}
    \label{eval_residual}
\end{minipage}
\end{table}

As shown in Table \ref{eval_voc_det}, LS-DFN improves R-FCN baseline model's mAP over 1.5\% with only $C' = 24$ dynamic features. This implies that the position-specific dynamic features are good supplement to the original feature space. And even though CoupleNets \cite{zhu2017couplenet} have already explicitly considered global information with large receptive fields, experimental results demonstrate that adding our LS-DFN block is still beneficial. \\

\begin{figure*}[t]
\centering
\includegraphics[width=0.85\linewidth,height=6cm,trim={1.5cm 1.5cm 1.5cm 1.5cm},clip]{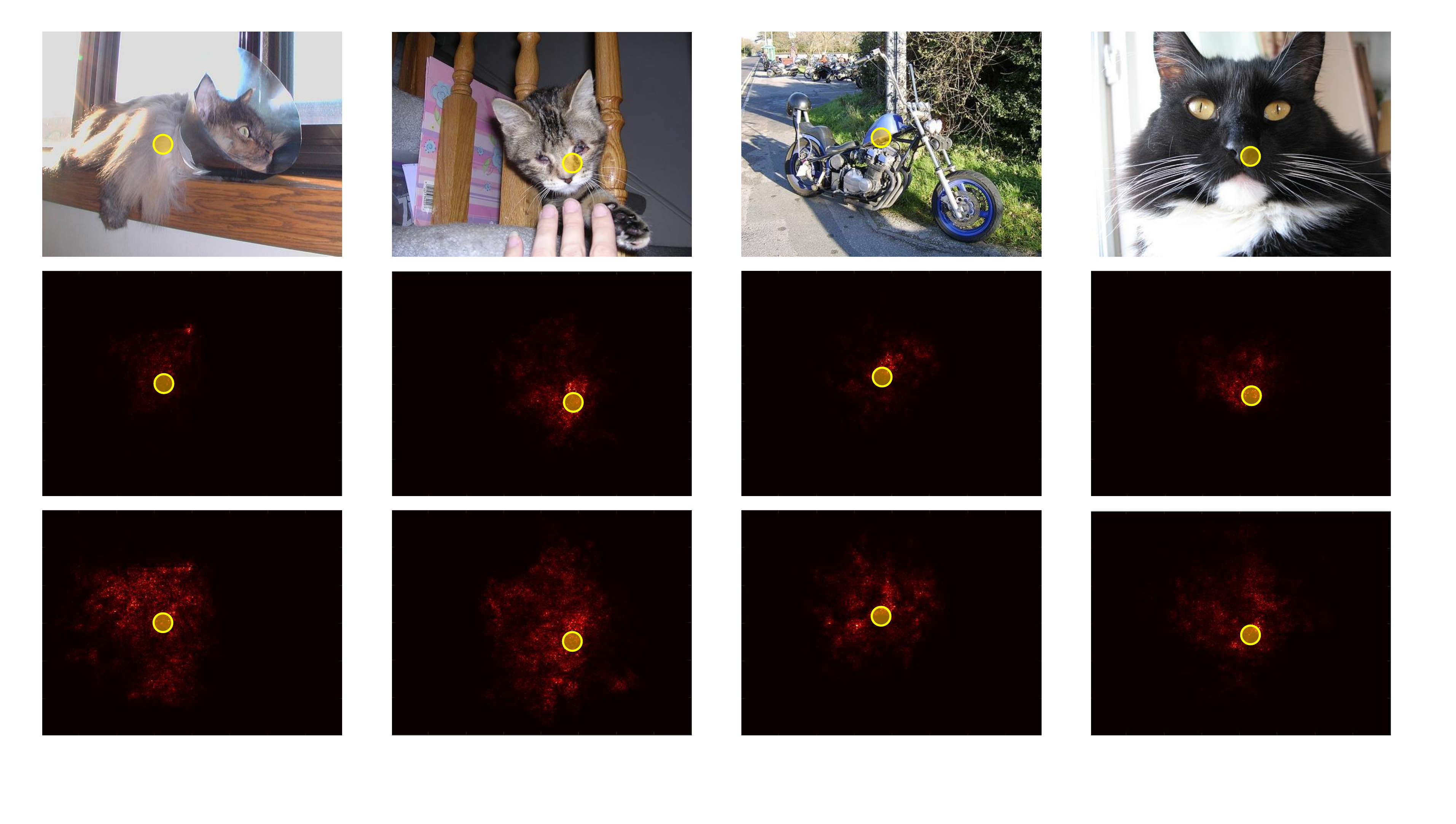}
\caption{Visualization on the effective receptive fields. The yellow circles denote the position on the objects. The first row presents input images. The second row contains the ERF figure from vanilla ResNet-50 model. The third row contains figures of the ERF with LS-DFNs. Best view in color.}
\label{fig:erf_fig}
\end{figure*}

\noindent\textbf{Evaluation on Effective Receptive Field.}
We evaluate the effective receptive fields (ERF) in the subsection. 
As illustrated in Fig. \ref{fig:erf_fig}, with ResNet-50 as backbone network, single additional LS-DFN layer provides much larger ERF than vanilla models thanks to the large sampling strategy. 
With larger ERFs, the networks can effectively observe larger region at each position thus can gather information and recognize objects more easily. 
Further, Table. \ref{eval_voc_det} experimentally verified the improvements on recognition abilities provided by our proposed LS-DFNs. \\

\noindent\textbf{Ablation Study on Sampling Size.}
We perform experiments to verify the advantages of applying more sampled regions in LS-DFN.

Table \ref{eval_sample} evaluates the effect of sampling in the neighbour regions. In simple DFN model \cite{de2016dynamic}, where $s=1$, though attention and residual learning strategy are adopted, the accuracy is still lower than R-FCN baseline (77.0\%). 
We argue the reason is that simple DFN model has limited receptive field. 
Besides, kernels at each position only receive gradients on the identical position which easily leads to overfitting.
With more sampled regions, we not only enlarge receptive field in feed-forward step, but also stabilize the gradients in back-propagation process. 
As shown in Table \ref{eval_sample}, when we take $3\times 3$ samples, the mAP score surpluses original R-FCN \cite{dai2016r} by 1.6\% and gets saturated with respect to $s$ when attention mechanism is applied. \\

\noindent\textbf{Ablation Study on Attention Mechanism.}
We verify the effectiveness of the attention mechanism in Table \ref{eval_attention} with different sample strides $\gamma$ and number of dynamic feature channels $C'$. 
In the experiments without attention mechanism, max pooling in channel dimension is adopted. 
We observe that, in nearly all cases, the attention mechanism helps improve mAP by more than 0.5\% in VOC2007 detection tasks. 
Especially as the number of dynamic feature channels $C'$ increases ($i.e.$ 32), the attention mechanism provides more benefits, increasing the mAP by 1\%, which indicates that the attention mechanism can further strengthen our LS-DFNs. \\

\noindent\textbf{Ablation Study on Residual Learning.}
We perform experiments to verify that with different numbers of dynamic feature channels, residual learning contributes a lot to the convergence of our LS-DFNs. 
As shown in Table \ref{eval_residual}, without utilizing residual learning, dynamic convolution models can hardly converge in real-world datasets. 
Even though they converge, the mAP is lower than expected.
When our LS-DFNs learn in a residual fashion, however, the mAP increase about 10\% on average. \\

\noindent\textbf{Runtime Analysis.}
Since the computation at each position and sampled regions can be done in a parallel fashion, the running time for the LS-DFN models could have potential of only slightly slower than two normal convolutional layers with kernel size $s^2$. 
\begin{figure*}[t]
\centering
\includegraphics[width=\textwidth]{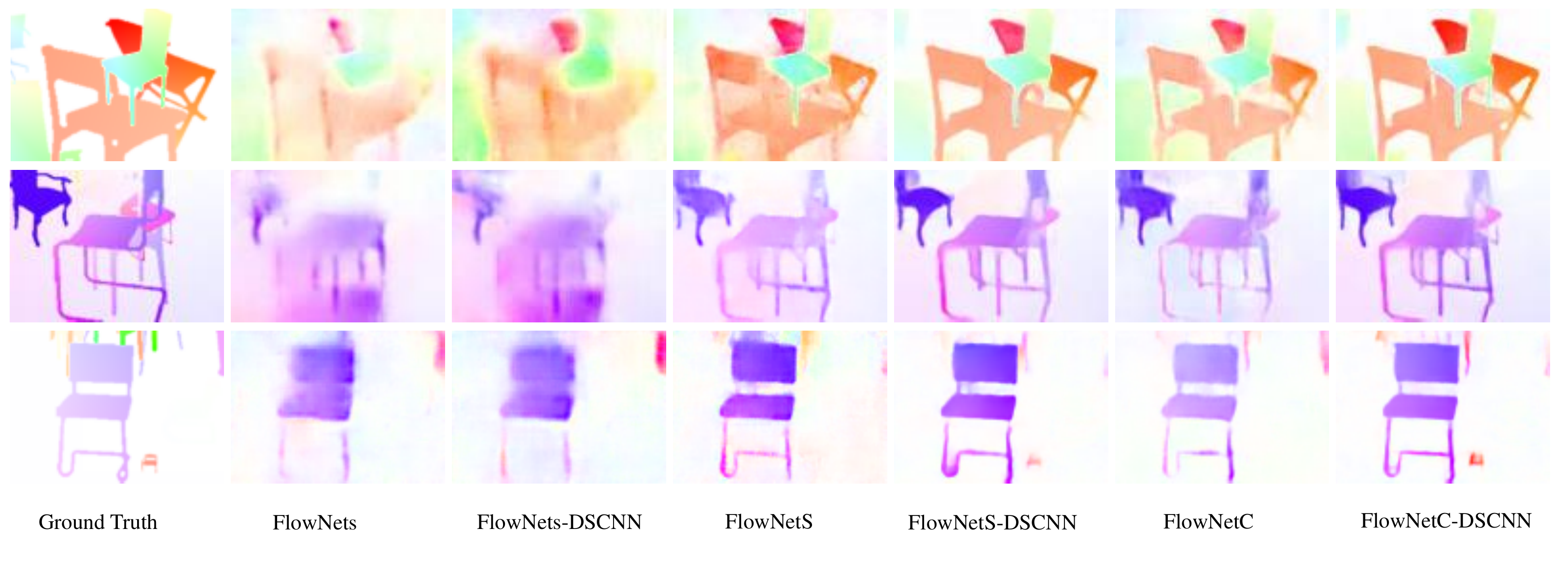}
\caption{Examples of Flow estimation on FlyingChairs dataset. The columns with LS-DFN denote the results of a LS-DFN added to the eir left columns. With LS-DFN, much sharper and more detailed optical flow can be estimated. }
\label{fig:opt_flow}
\end{figure*}

\subsection{Semantic Segmentation}
\begin{table*}[!t]
	\centering
	\scriptsize{
		\begin{tabular}{L{3.6cm}|C{0.7cm}C{0.7cm}C{0.6cm}C{0.8cm}C{0.8cm}C{0.8cm}C{0.6cm}C{0.6cm}C{0.6cm}C{0.6cm}C{0.6cm}}
			\hline
			Methods &  {bg} &  {aero} &   {bike} &  {bird} &  \textbf{boat} &  {bottle} &  {bus} &  {car}  &  {cat} & \textbf{chair}&  {cow}  \\\hline
			DeepLabV2 + CRF  & -      & 92.6	 & \textbf{60.4} & \textbf{ 91.6}	&  63.4	 &  76.3  &  95.0	&  88.4	&  92.6&    	32.7	& 88.5\\
			... $w/o \ atrous$ +LS-DFN  & 95.3 & 92.3  &  57.2  & 91.1 & 68.8 & 76.8 & 95.0 & 88.8 & 92.1 & 35.0  & 88.5 \\ 
			... + SegAware \cite{harley2017segmentation} & {95.3} & {92.4} &  {58.5}&  {91.3} & {65.6} & {76.8} &  {95.0} &  {88.7} &  {92.1} &  {34.7} &  {88.5} \\
			... + LS-DFN$^\dag   $    & \textbf{95.5} & \textbf{94.0} & {58.5} & {91.3}  & \textbf{69.2} & \textbf{78.2} &  {95.4} & \textbf{89.6} & \textbf{92.9} & {38.4} & {89.9} \\
		\end{tabular} 
		\begin{tabular}{L{3.6cm}|C{0.7cm}C{0.7cm}C{0.6cm}C{0.8cm}C{0.8cm}C{0.8cm}C{0.6cm}C{0.6cm}C{0.6cm}C{0.6cm}C{0.6cm}}		
			\hline
			Methods&  {table} &  {dog} &  {horse} &  {motor} &  {person} &  {plant} &  {sheep} &  \textbf{sofa}  &  {train} &  {tv} & {{\textbf{all}}}  \\\hline	
			DeepLabV2 + CRF                                &   67.6  &	89.6  &	 92.1	&   87.0 &	\textbf{87.4}  &	 63.3	& 88.3 &	60.0 &	86.8 &	74.5  & 79.7\\	
			... $w/o \ atrous$ + LS-DFN  &  68.7 & 89.0  &  92.2  & \textbf{87.1} & 87.1 & 63.3& 88.4 & 64.1  & 88.0  & 74.8& 80.4 \\ 	
			... + SegAware \cite{harley2017segmentation} &  {68.7} & {89.0} &  {92.2} &  {87.0} & {87.1} & \textbf{63.4} & {88.4} & {60.9} & {86.3} & {74.9} &{79.8}\\

			... + LS-DFN$^\dag$     & \textbf{70.2} & {90.8} &  {93.1} &  {87.0} &  \textbf{87.4} &  \textbf{63.4} &  \textbf{89.5} &  {64.9} &  \textbf{88.9} &  \textbf{75.8} & \textbf{81.1} \\

			\hline
			
		\end{tabular}
	}
	\caption{Performance comparison on the PASCAL VOC 2012 semantic segmentation test set. The average IoU (\%) for each class and the overall IoU is reported. $^\dag$\url{http://host.robots.ox.ac.uk:8080/anonymous/5SYVME.html}}
	\label{tab:voc_seg}
\end{table*}

We adopt the DeepLabV2 with CRF as the baseline model. The added LS-DFN layer receives input features from res5b layer in ResNet-101 and its output features are
concatenated to the res5c layer. For hyperparameters, we adopt $C' = 24$, $s = 5$, $\gamma = 3$, $k=3$ and a $1\times 1$ 256-channel post-conv layer with shared weights at all three input scales. Following SegAware \cite{harley2017segmentation}, we initialize the network with ImageNet model, then train on COCO trainval sets, and finetune on the augmented PASCAL images. 

We report the segmentation results in Table. \ref{tab:voc_seg}. Our model achieves 81.2\% overall IoU accuracy which is 1.4\% superior to SegAware DeepLab-V2. Furthermore, the results on large objects like boat and sofa\footnote{We observe most boat and sofa instances occupy large area in images in PASCAL VOC test set.} are significantly improved ($i.e.$ 3.6\% in boat and 4.2\% in sofa). The reason is that the LS-DFN layer is capable of significantly enlarging the effective receptive fields (ERF) so that the pixels inside the objects can utilize a much wider context, which is important since the visual clues of determining the correct categories for the pixels can be far away from the pixels themselves. 

It's worth noting that the performance of the chair category is also significantly improved thanks to the reduced false positive classification where many pixels in sofa instances are originally classified as chairs'.

We use $w/o \ atrous$+LS-DFN to denote the DeepLabV2 model where all the dilated convolutions are replaced by LS-DFN block in Table. \ref{tab:voc_seg}. In particular, the different dilation rates 6, 12, 18, 24 are replaced by sample strides $\gamma = 2, 4, 6, 8$ in the LS-DFN layers. And all branches are implemented as single conv layers with $k=3$,  $s=5$, $C'=21$ for classification. Compared with original DeepLabV2 model, we observe a considerable improvement ($i.e.$ from 79.7 \% to 80.4\%) indicating that the LS-DFN layers are able to better model the contextual information within the large receptive fields thanks to the dynamic sampling kernels.

\subsection{Optical Flow Estimation}
We perform experiments on optical flow estimation using the FlyingChairs dataset \cite{DFIB15}. 
This dataset is a synthetic one with optical flow ground truth and widely used in deep learning methods to learn the motion information. 
It consists of 22872 image pairs and corresponding flow fields. 
In experiments we use FlowNets(S) and FlowNetC \cite{ilg2016flownet} as our baseline models, though other complicated models are also applicable. 
All of the baseline models are fully-convolutional networks which firstly downsample input image pairs to learn semantic features then upsample the features to estimate optical flow.

In experiments, our LS-DFN model is inserted in a relative shallower layer to produce sharper optical flow images. 
Specifically, we adopt the third conv layer, where image pairs are merged into a single branch volume in FlowNetC model. 
We also use skip-connection to connect the LS-DFN outputs to the corresponding upsampling layer. 
In order to capture large displacement, we apply more samples in our LS-DFN layer. 
Concretely, we use $7\times7$ or $9\times9$ samples with a sample stride of 2 in our experiments. 
We follow similar training process in \cite{dosovitskiy2015flownet} for fair comparison\footnote{We use 300k iterations with double batchsize}.
As shown in Fig. \ref{fig:opt_flow}, our LS-DFN models are able to output sharper and more accurate optical flow. We argue this is due to the large receptive fields and dynamic position-specific kernels. 
Since each position estimates optical flow with its own kernels, our LS-DFN can better identify the contours of the moving objects. 
\begin{figtab}[!t]
\begin{minipage}[!t]{0.45\textwidth} 
\centering
\includegraphics[width=\linewidth,height=\linewidth,trim={0cm 5cm 0cm 6cm},clip]{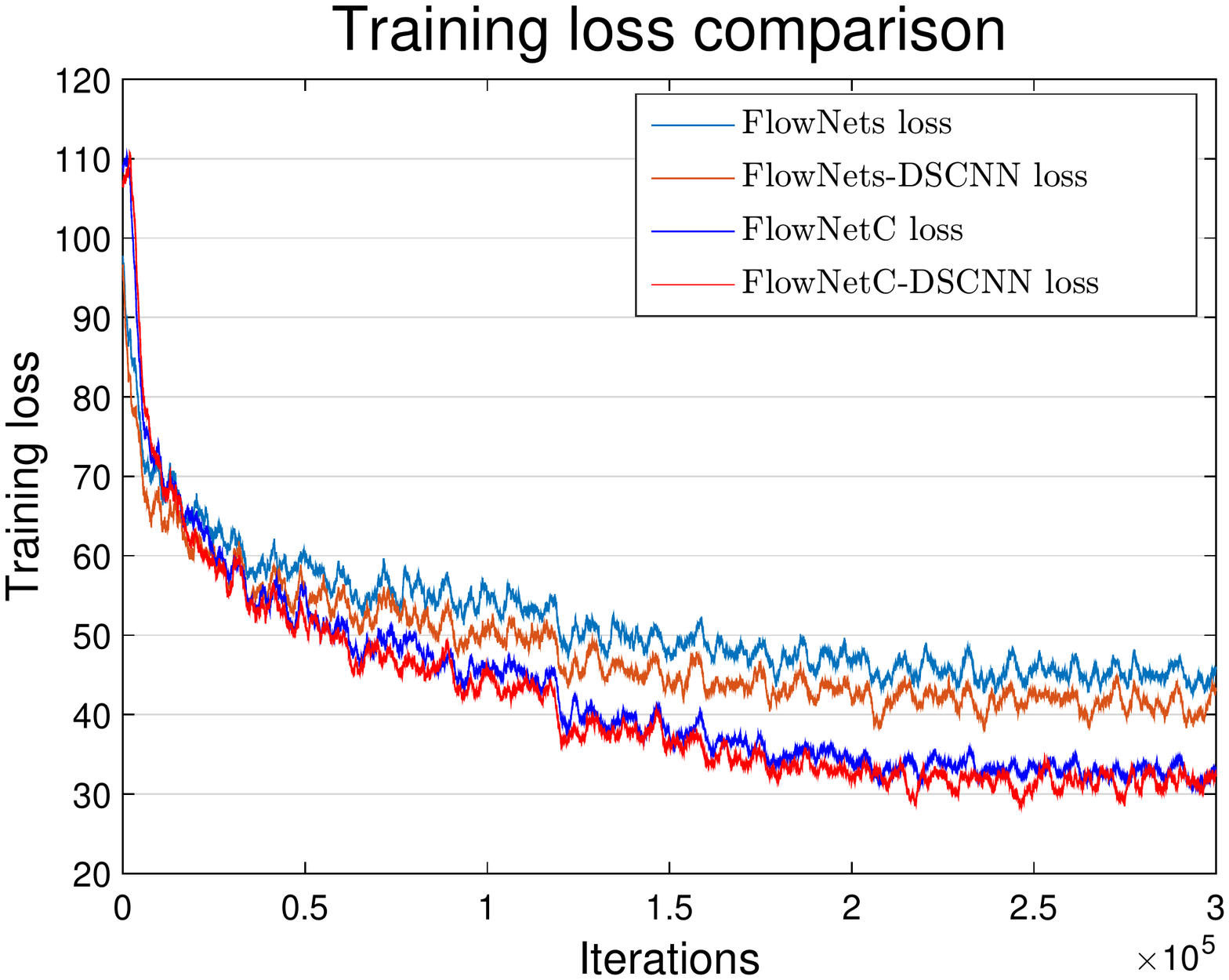}
\figcaption{Training loss of flow estimation. We use moving average with window size of 2k iterations when plotting the loss curve.}
\label{fig:flownet_loss}
\end{minipage}%
  \qquad
\begin{minipage}[!t]{0.45\textwidth} 
\centering
\begin{tabular}{L{3.9cm}|C{0.8cm}|C{0.8cm}}
\hline
	model & aEPE &  Time\\ \hline\hline
	Spynet \cite{ranjan2016optical} & 2.63 & -\\ 
	EpicFlow \cite{revaud2015epicflow} & 2.94 & -  \\ 
	DeepFlow \cite{weinzaepfel2013deepflow} & 3.53 & -\\ 
	PWC-Net \cite{sun2017pwc} & 2.26 & - \\ \hline
	FlowNets \cite{ilg2016flownet} & 3.67 & 6ms\\ 
	FlowNets+LS-DFN, $s=7$ & \textbf{2.88} &23ms\\ \hline
	FlowNetS \cite{ilg2016flownet}  & 2.78 & 16ms\\ 
	FlowNetS+SegAware \cite{harley2017segmentation} & 2.36& - \\ 
	FlowNetS+LS-DFN,$s=7 $ & \textbf{2.34} &34ms\\ \hline
	FlowNetC \cite{ilg2016flownet}  & 2.19 & 25ms\\ 
	FlowNetC+LS-DFN, $s=7$ & 2.11 &43ms\\ 
	FlowNetC+LS-DFN, $s=9$ &\textbf{2.06}&51ms\\ \hline
\end{tabular}
\tabcaption{aEPE and running time evaluation of optical flow estimation.}
\label{tab:eval_flow}
\end{minipage} 
\end{figtab}

As shown in Fig. \ref{fig:flownet_loss}, LS-DFN model successfully relaxes the constraint of sharing kernels spatially and converges to a lower training loss in both FlowNets and FlowNetC models. That further indicates the advantages of local gradients when doing dense prediction tasks.

We use average End-Point-Error (aEPE) to quantitatively measure the performance of the optical flow estimation. 
As shown in Table \ref{tab:eval_flow}, with a single LS-DFN layer added, the aEPEs decrease in all baseline models by a large margin. 
In FlowNets model, aEPE decreases by 0.79 which demonstrates the increased learning capacity and robustness of our LS-DFN model. 
Even though SegAware attention model \cite{harley2017segmentation} explicitly takes advantage of boundary information which requires additional training data, our LS-DFN can still slightly outperforms them using FlowNetS as baseline model. With $s=9$ and $\gamma=2$, we have approximately $40$ times larger receptive fields which allow the FlowNet models to easily capture large displacements in flow estimation task in FlyingChairs dataset.  

\section{Conclusion}
This work introduces Dynamic Filtering with Large Sampling Field (LS-DFN) to learn dynamic position-specific kernels and takes advantage of very large receptive fields and local gradients. 
Thanks to the large ERF in a single layer, LS-DFNs have better performance in most general tasks. 
With local gradients and dynamic kernels, LS-DFNs are able to produce much sharper output features, which is beneficial especially in dense prediction tasks such as optical flow estimation. \\

\noindent\textbf{Acknowledgements.} Supported by National Key R\&D Program of China under contract No.2017YFB1002202, Projects of International Cooperation and Exchanges NSFC with No. 61620106005, National Science Fund for Distinguished Young Scholars with No. 61325003, Beijing Municipal Science \& Technology Commission Z181100008918014 and Tsinghua University Initiative Scientific Research Program.

\clearpage

\bibliographystyle{splncs04}
\bibliography{egbib}

\begin{thebibliography}{10}
\providecommand{\url}[1]{\texttt{#1}}
\providecommand{\urlprefix}{URL }
\providecommand{\doi}[1]{https://doi.org/#1}

\bibitem{chen2016deeplab}
Chen, L.C., Papandreou, G., Kokkinos, I., Murphy, K., Yuille, A.L.: Deeplab:
  Semantic image segmentation with deep convolutional nets, atrous convolution,
  and fully connected crfs. arXiv preprint arXiv:1606.00915  (2016)

\bibitem{dai2016instance}
Dai, J., He, K., Li, Y., Ren, S., Sun, J.: Instance-sensitive fully
  convolutional networks. In: European Conference on Computer Vision. pp.
  534--549. Springer (2016)

\bibitem{dai2016r}
Dai, J., Li, Y., He, K., Sun, J.: R-fcn: Object detection via region-based
  fully convolutional networks. In: Advances in neural information processing
  systems. pp. 379--387 (2016)

\bibitem{dai2017deformable}
Dai, J., Qi, H., Xiong, Y., Li, Y., Zhang, G., Hu, H., Wei, Y.: Deformable
  convolutional networks. arXiv preprint arXiv:1703.06211  (2017)

\bibitem{de2016dynamic}
De~Brabandere, B., Jia, X., Tuytelaars, T., Van~Gool, L.: Dynamic filter
  networks. In: Neural Information Processing Systems (NIPS) (2016)

\bibitem{DFIB15}
Dosovitskiy, A., Fischer, P., Ilg, E., H{\"a}usser, P., Haz{\i}rba{\c{s}}, C.,
  Golkov, V., v.d. Smagt, P., Cremers, D., Brox, T.: Flownet: Learning optical
  flow with convolutional networks. In: IEEE International Conference on
  Computer Vision (ICCV) (2015),
  \url{http://lmb.informatik.uni-freiburg.de//Publications/2015/DFIB15}

\bibitem{dosovitskiy2015flownet}
Dosovitskiy, A., Fischer, P., Ilg, E., Hausser, P., Hazirbas, C., Golkov, V.,
  van~der Smagt, P., Cremers, D., Brox, T.: Flownet: Learning optical flow with
  convolutional networks. In: Proceedings of the IEEE International Conference
  on Computer Vision. pp. 2758--2766 (2015)

\bibitem{Everingham10}
Everingham, M., Van~Gool, L., Williams, C.K.I., Winn, J., Zisserman, A.: The
  pascal visual object classes (voc) challenge. International Journal of
  Computer Vision  \textbf{88}(2),  303--338 (Jun 2010)

\bibitem{Girshick_2015_ICCV}
Girshick, R.: Fast r-cnn. In: The IEEE International Conference on Computer
  Vision (ICCV) (December 2015)

\bibitem{harley2017segmentation}
Harley, A.W., Derpanis, K.G., Kokkinos, I.: Segmentation-aware convolutional
  networks using local attention masks. arXiv preprint arXiv:1708.04607  (2017)

\bibitem{He_2016_CVPR}
He, K., Zhang, X., Ren, S., Sun, J.: Deep residual learning for image
  recognition. In: The IEEE Conference on Computer Vision and Pattern
  Recognition (CVPR) (June 2016)

\bibitem{Howard2017MobileNetsEC}
Howard, A.G., Zhu, M., Chen, B., Kalenichenko, D., Wang, W., Weyand, T.,
  Andreetto, M., Adam, H.: Mobilenets: Efficient convolutional neural networks
  for mobile vision applications. CoRR  \textbf{abs/1704.04861} (2017)

\bibitem{ilg2016flownet}
Ilg, E., Mayer, N., Saikia, T., Keuper, M., Dosovitskiy, A., Brox, T.: Flownet
  2.0: Evolution of optical flow estimation with deep networks. arXiv preprint
  arXiv:1612.01925  (2016)

\bibitem{kim2016multimodal}
Kim, J.H., Lee, S.W., Kwak, D., Heo, M.O., Kim, J., Ha, J.W., Zhang, B.T.:
  Multimodal residual learning for visual qa. In: Advances in Neural
  Information Processing Systems. pp. 361--369 (2016)

\bibitem{NIPS2012_4824}
Krizhevsky, A., Sutskever, I., Hinton, G.E.: Imagenet classification with deep
  convolutional neural networks. In: Pereira, F., Burges, C.J.C., Bottou, L.,
  Weinberger, K.Q. (eds.) Advances in Neural Information Processing Systems 25,
  pp. 1097--1105. Curran Associates, Inc. (2012),
  \url{http://papers.nips.cc/paper/4824-imagenet-classification-with-deep-convolutional-neural-networks.pdf}

\bibitem{li2016fully}
Li, Y., Qi, H., Dai, J., Ji, X., Wei, Y.: Fully convolutional instance-aware
  semantic segmentation. arXiv preprint arXiv:1611.07709  (2016)

\bibitem{lin2016feature}
Lin, T.Y., Doll{\'a}r, P., Girshick, R., He, K., Hariharan, B., Belongie, S.:
  Feature pyramid networks for object detection. arXiv preprint
  arXiv:1612.03144  (2016)

\bibitem{Long_2015_CVPR}
Long, J., Shelhamer, E., Darrell, T.: Fully convolutional networks for semantic
  segmentation. In: The IEEE Conference on Computer Vision and Pattern
  Recognition (CVPR) (June 2015)

\bibitem{long2016unsupervised}
Long, M., Zhu, H., Wang, J., Jordan, M.I.: Unsupervised domain adaptation with
  residual transfer networks. In: Advances in Neural Information Processing
  Systems. pp. 136--144 (2016)

\bibitem{Luo2016UnderstandingTE}
Luo, W., Li, Y., Urtasun, R., Zemel, R.S.: Understanding the effective
  receptive field in deep convolutional neural networks. In: NIPS (2016)

\bibitem{ranjan2016optical}
Ranjan, A., Black, M.J.: Optical flow estimation using a spatial pyramid
  network. arXiv preprint arXiv:1611.00850  (2016)

\bibitem{ren2015faster}
Ren, S., He, K., Girshick, R., Sun, J.: Faster r-cnn: Towards real-time object
  detection with region proposal networks. In: Advances in neural information
  processing systems. pp. 91--99 (2015)

\bibitem{revaud2015epicflow}
Revaud, J., Weinzaepfel, P., Harchaoui, Z., Schmid, C.: Epicflow:
  Edge-preserving interpolation of correspondences for optical flow. In:
  Proceedings of the IEEE Conference on Computer Vision and Pattern
  Recognition. pp. 1164--1172 (2015)

\bibitem{sharma2015action}
Sharma, S., Kiros, R., Salakhutdinov, R.: Action recognition using visual
  attention. arXiv preprint arXiv:1511.04119  (2015)

\bibitem{simonyan2014very}
Simonyan, K., Zisserman, A.: Very deep convolutional networks for large-scale
  image recognition. arXiv preprint arXiv:1409.1556  (2014)

\bibitem{srivastava2015unsupervised}
Srivastava, N., Mansimov, E., Salakhudinov, R.: Unsupervised learning of video
  representations using lstms. In: International Conference on Machine
  Learning. pp. 843--852 (2015)

\bibitem{sun2017pwc}
Sun, D., Yang, X., Liu, M.Y., Kautz, J.: Pwc-net: Cnns for optical flow using
  pyramid, warping, and cost volume. arXiv preprint arXiv:1709.02371  (2017)

\bibitem{szegedy2015going}
Szegedy, C., Liu, W., Jia, Y., Sermanet, P., Reed, S., Anguelov, D., Erhan, D.,
  Vanhoucke, V., Rabinovich, A.: Going deeper with convolutions. In:
  Proceedings of the IEEE conference on computer vision and pattern
  recognition. pp.~1--9 (2015)

\bibitem{wang2017residual}
Wang, F., Jiang, M., Qian, C., Yang, S., Li, C., Zhang, H., Wang, X., Tang, X.:
  Residual attention network for image classification. arXiv preprint
  arXiv:1704.06904  (2017)

\bibitem{weinzaepfel2013deepflow}
Weinzaepfel, P., Revaud, J., Harchaoui, Z., Schmid, C.: Deepflow: Large
  displacement optical flow with deep matching. In: Proceedings of the IEEE
  International Conference on Computer Vision. pp. 1385--1392 (2013)

\bibitem{wu2016action}
Wu, J., Wang, G., Yang, W., Ji, X.: Action recognition with joint attention on
  multi-level deep features. arXiv preprint arXiv:1607.02556  (2016)

\bibitem{xu2015show}
Xu, K., Ba, J., Kiros, R., Cho, K., Courville, A., Salakhudinov, R., Zemel, R.,
  Bengio, Y.: Show, attend and tell: Neural image caption generation with
  visual attention. In: International Conference on Machine Learning. pp.
  2048--2057 (2015)

\bibitem{zhu2017couplenet}
Zhu, Y., Zhao, C., Wang, J., Zhao, X., Wu, Y., Lu, H.: Couplenet: Coupling
  global structure with local parts for object detection

\end{thebibliography}
\end{document}